\documentclass{article}

\usepackage{booktabs}
\usepackage{pifont}

\usepackage[preprint]{neurips_2023}

\usepackage[utf8]{inputenc} 
\usepackage[T1]{fontenc}    
\usepackage{hyperref}       
\usepackage{url}            
\usepackage{booktabs}       
\usepackage{amsfonts}       
\usepackage{nicefrac}       
\usepackage{microtype}      
\usepackage{xcolor}         
\usepackage{graphicx}
\usepackage{amsmath}
\usepackage{amssymb}
\usepackage{amsfonts} 
\usepackage{amsmath}
\usepackage{algorithm}
\usepackage{algpseudocode}
\usepackage{subcaption}
\usepackage{diagbox}
\usepackage{booktabs}
\usepackage{multirow}
\usepackage{pgfplots}
\usepackage{adjustbox}

\title{Improving 3D Few-Shot Segmentation with Inference-Time Pseudo-Labeling}

\author{%
  Mohammad~Mozafari\\
  \texttt{mohammadmozafari@sharif.edu} \\
  \And
  Hosein~Hasani \\  \texttt{hosein.hasani@sharif.edu} \\
  \AND
  Reza Vahidimajd \\
  \texttt{reza.vahidimajd202@sharif.edu} \\
  \And
  Mohamadreza Fereydooni \\
  \texttt{fereydooni@sharif.edu} \\
  \And
  Mahdieh Soleymani Baghshah \\
  \texttt{soleymani@sharif.edu}
  \vspace{3mm}\\
  Department of Computer Engineering \\
  Sharif University of Technology \\
  Tehran, Iran \\
}

\begin{document}

\maketitle

\begin{abstract}
  In recent years, few-shot segmentation (FSS) models have emerged as a promising approach in medical imaging analysis, offering remarkable adaptability to segment novel classes with limited annotated data. Existing approaches to few-shot segmentation have often overlooked the potential of the query itself, failing to fully utilize the valuable information it contains. However, treating the query as unlabeled data provides an opportunity to enhance prediction accuracy. Specifically in the domain of medical imaging, the volumetric structure of queries offers a considerable source of valuable information that can be used to improve the target slice segmentation. In this work, we present a novel strategy to efficiently leverage the intrinsic information of the query sample for final segmentation during inference.  First, we use the support slices from a reference volume to generate an initial segmentation score for the query slices through a prototypical approach. Subsequently, we apply a confidence-aware pseudo-labeling procedure to transfer the most informative parts of query slices to the support set. The final prediction is performed based on the new expanded support set, enabling the prediction of a more accurate segmentation mask for the query volume. Extensive experiments show that the proposed method can effectively boost performance across diverse settings and datasets.
\end{abstract}

\vspace{3mm}
\section{Introduction}

Image segmentation is a primary problem in the medical imaging field, crucial for tasks like disease diagnosis and treatment planning \cite{tsochatzidis2021integrating,chen2021deep}. Deep learning methods have sped up progress in medical image analysis, particularly in automated image segmentation. However, their effectiveness heavily relies on extensive annotated data, which is often scarce, especially for 3D volumetric images that require distinct annotations for the 2D slices of each 3D scan.
Supervised deep learning methods struggle with generalizing to novel classes, necessitating innovative segmentation strategies for limited annotated data.

Few-shot segmentation offers a promising way of addressing data scarcity challenges by leveraging meta-learning principles. 
The task involves training a model capable of segmenting novel classes in previously unseen images using only a limited number of annotated examples. For this purpose, the model utilizes a few labeled examples (referred to as the support set) to obtain a distinct representation for each class and leverages the extracted information to perform segmentation on the unlabeled images (referred to as the query set).
Recently, prototypical methods have achieved the best results in this domain. Based on the ProtoNet \cite{snell2017prototypical}, many works focus on extracting prototypes \cite{zhang2020sg,wang2019panet,cao2022prototype,zhang2022mask,lang2022beyond,yang2020prototype,liu2022learning,zhang2022feature,okazawa2022interclass,wang2022adaptive}   for each class using the support set and then using similarity metrics to segment the query images. For instance, PANet \cite{wang2019panet} introduced a straightforward approach by extracting a single prototype for each class and using these prototypes for segmentation.

In medical imaging, where data scarcity is common, few-shot segmentation efficiently handles rare or novel structures within MRI and CT datasets with minimal labeled examples. This approach facilitates segmentation, enhances performance, and reduces the need for extensive data labeling, especially in volumetric medical data. In the context of medical imaging, SE-Net \cite{roy2020squeeze} was the first few-shot segmentation method. SE-Net operates through a dual-branch structure comprising a conditioner and a segmenter, utilizing squeeze and excite blocks \cite{hu2018squeeze} to segment query images based on provided labeled support sets. On the other hand, \cite{ouyang2020self} adopts a prototypical methodology for the few-shot segmentation of medical images. To preserve local information, SSLALPNet \cite{ouyang2020self} extracts local prototypes computed on regular grids, while also addressing the scarcity of annotated data through self-supervised training employing superpixels.

While few-shot learning has become prevalent by addressing data scarcity issues, specifically in the domain of medical imaging, it is not without its limitations and challenges. One significant challenge arises from the difficulty of training few-shot segmentation models for medical imaging, which often requires large meta-training datasets with numerous annotated classes to prevent overfitting. Moreover, the intrinsic disparity between limited and fixed support images and arbitrary query images can lead to failures in capturing underlying appearance variations of target classes, exacerbated by data scarcity and diversity issues inherent in medical few-shot learning.
However, in the few-shot segmentation as opposed to few-shot classification, a rich source of unlabeled data is available since the 2D (and more prominently the 3D) query data has numerous unlabeled pixels (or voxels). This intrinsic characteristic offers the potential to significantly boost segmentation accuracy.
In semi-supervised learning \cite{zhu2022introduction}, unlabeled samples are combined with labeled samples to facilitate the learning process. Integrating semi-supervised learning techniques into few-shot learning has garnered considerable interest as it can substantially enhance data efficiency \cite{ren2018metalearning,lin2023contrastive,lazarou2021iterative,wei2022embarrassingly,wang2020instance,li2019learning}. Surprisingly, this avenue has not received attention in the realm of few-shot segmentation of medical images.

In this work, we aim to exploit the untapped potential of query data within the few-shot segmentation framework, akin to semi-supervised settings. We present a novel strategy to effectively exploit the valuable information embedded within the query samples to enhance the accuracy of segmentation during inference. In the first stage of our method, an initial segmentation score for the query slices is obtained by using annotated support slices like the traditional prototypical approach. Based on initial query segmentation scores, a confidence-aware pseudo-labeling technique is designed to transfer key informative segments from the query slices to enrich the support set. Finally, we use the augmented support set to segment target query slices. Through extensive experimentation, our proposed method has demonstrated its effectiveness in enhancing FSS performance across diverse settings and datasets.

\vspace{3mm}
\section{Method}

In this section, we begin by reviewing the problem formulation for few-shot segmentation in medical imaging. Subsequently, we introduce our method, which focuses on effectively leveraging query data to enhance segmentation accuracy during inference. The methodology encompasses key components such as support slice utilization, pseudo-labeling techniques, and an expanded support set to improve segmentation outcomes. 

\vspace{2mm}
\subsection{Problem Setup}

In the context of few-shot medical image segmentation, the problem is formulated as follows. Let the source training dataset be denoted as $\mathcal{D}_{src} = \{(x_i, y_i(c)\}_{i=1}^{N_{src}}$, where $x_i$ represents the medical image and $y_i^c$ is the corresponding binary mask with semantic label $c$. Similarly, the target dataset containing novel classes is denoted as $\mathcal{D}_{trg} = \{(x_i, y_i(c))\}_{i=1}^{N_{trg}}$.
Here, $N_{src}$ and $N_{trg}$ denote the number of samples in the training and testing datasets, respectively. The sets of source and target classes are denoted as
$\mathcal{C}_{src} = \{c|c \in \mathcal{D}_{src}\}$ and $\mathcal{C}_{trg} = \{c|c \in \mathcal{D}_{trg}\}$, where $\mathcal{C}_{src} \cap \mathcal{C}_{trg} = \varnothing$.

In the few-shot learning paradigm, the model $f_\theta$ is trained on $\mathcal{D}_{src}$ with the objective of predicting an unseen class $c \in \mathcal{C}_{trg}$ during the meta-testing phase, given only a few support examples from $\mathcal{D}_{trg}$. Specifically, the few-shot segmentation model is trained to operate in an $N$-way $K$-shot setting, where $N$ represents the number of semantic classes to be segmented, and $K$ is the number of examples available for each class during the query phase.

During training, an episodic training strategy is employed, simulating the conditions of the final testing phase where only $K$ examples for each class are available. Each episode consists of two sets of data randomly sampled from $\mathcal{D}_{src}$, a support set $\mathcal{S} = \{\mathcal{S}^c\}_{c=1}^N$ with $\mathcal{S}^c = \{(x^{s,c}_k,y^{s,c}_k\}_{k=1}^{K}$ representing few-shot training samples and a query set $\mathcal{Q}$, representing the unseen classes to be segmented. The model is trained to distill knowledge about a semantic class from the support set and apply this knowledge to segment the query set during the testing phase. During inference, only the support images and their corresponding labels are provided, and the model performs segmentation on the query images.

The goal is to develop a few-shot segmentation model $f_\theta$ that can generalize effectively to novel classes during the testing phase, addressing the challenges posed by limited annotated data in the medical imaging domain.

\vspace{2mm}
\subsection{Confidence-Aware Semi-Supervised FSS}

To address the challenges posed by limited annotated data in few-shot segmentation for medical imaging, we introduce a novel procedure designed to efficiently exploit the intrinsic information contained within query samples for precise segmentation during inference.
The main steps are outlined as follows.

To initiate the segmentation process, we leverage the support slices from a reference volume to generate an initial segmentation score for the corresponding query slices. Following a prototypical approach, we calculate the prototype vectors for each class based on the annotated support slices. Using feature embeddings of support set
$f_\theta(x_k^c) \in \mathbb{R}^{H\times W \times Z}$, the prototype for class $c$ is computed through the masked average pooling:
\begin{equation}
    \mathcal{P}^{s,c} = \frac{1}{K} \sum_{k=1}^{K} \frac{\sum\limits_{h,w} f_\theta(x^{s,c}_k)(h,w) \cdot y^{s,c}_{k}(h,w)}{\sum\limits_{h, w} y^{s,c}_{k}(h,w)} \in \mathbb{R}^{Z}    
\end{equation}

Here, the indices $(h, w)$ reference pixels on the feature map, and $y^c(h,w)$ denotes spatial locations within the binary mask for class $c$.
The prototype vectors are then used to obtain the initial segmentation score for the query slices.
Probabilities of all classes are obtained by applying a softmax function to query distances.
Here we use cosine distance to calculate distances between each query embedding and support prototypes. Specifically, for each pixel at location $(h, w)$ within the query feature map $f_\theta(q)$, the cosine distance and softmax probabilities are expressed as:
\begin{equation}
    d(f_\theta(x^q_k)(h,w),\mathcal{P}^{s,c})=\frac{f_\theta(x^q_k)(h,w)\cdot \mathcal{P}^{s,c}}{\Vert f_\theta(x^q_k)(h,w)\Vert_2\Vert \mathcal{P}^{s,c}\Vert_2}
\end{equation}
\begin{equation}
p^{q,c}(h,w) = \frac{\exp\left(-d(f_\theta(x^q_k)(h,w), \mathcal{P}^{s,c})\right)}{\sum\limits_{n=1}^{N} \exp\left(-d(f_\theta(x^q_k)(h,w), \mathcal{P}^{s,n})\right)}
\end{equation}

In the above equations, $d(., .)$ represents the cosine distance between two vectors, and $p^{q,c}(h,w)$ denotes the probability that the query pixel at location $(h, w)$ belongs to class $c$.
In the second step, we employ a confidence-aware pseudo-labeling procedure. This step is crucial for transferring key informative segments from the query slices to enrich the support set.
By considering a confidence level $\gamma$ for the initial query segmentation probabilities, we identify the most informative regions within the query slices.
Specifically, our model predicts a segmentation probability map for each pixel in the query images, indicating the probability of each pixel belonging to different classes.
Subsequently, these informative segments are selected to be incorporated into the support set with pseudo-labels $\hat{y}^q$:
\begin{equation}
    \hat{y}^q(h,w) = \underset{c}{\operatorname*{argmax}} p^{q,c}(h,w)
\end{equation}

This pseudo-labeling procedure is applied to the $M$ consecutive slices of the same query volume. More confident pixels of each slice are selected to extract prototypes of query slices.
Regions in the query images with probability scores exceeding a threshold $\gamma$, will be used to form the query prototypes:

\begin{equation}
    \mathcal{P}^{q,c} = \frac{1}{M} \sum_{m=1}^{M} \sum\limits_{\substack{h,w \\ p^{q,c}_m(h,w) \geq \gamma^c}} \frac{ f_\theta(x^{q}_m)(h,w) \cdot \hat{y}^{q,c}_{m}(h,w)}{ \hat{y}^{q,c}_{m}(h,w)} \in \mathbb{R}^{Z}    
\end{equation}

\begin{figure}[!htb]
  \centering
  \includegraphics[width=0.99\linewidth]{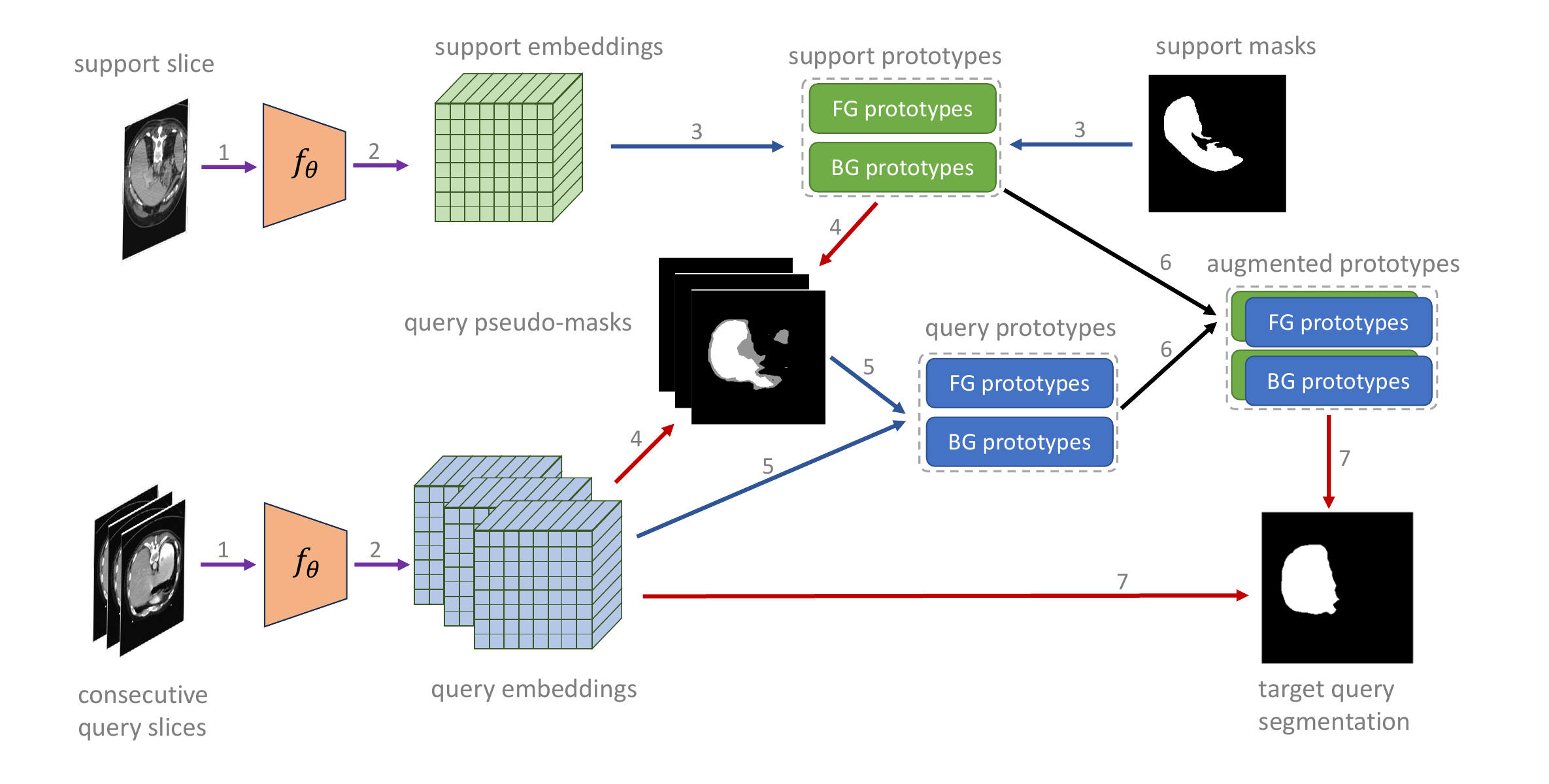}
   \caption{
    Overview of the proposed method.
    The workflow begins with the extraction of embeddings from both the support and query slices using a shared feature extractor (Steps 1 and 2). In Step 3, support prototypes are generated from the support embeddings and corresponding ground truth masks.
    Step 4 calculates pseudo-masks for the query by measuring the distance between the support prototypes and the query embeddings. Next, in Step 5, query prototypes are generated from the query embeddings and pseudo-masks, and these are combined with the support prototypes in Step 6. Finally, in Step 7, the segmentation of the query slice is performed by using the augmented prototypes along with the query embeddings.
    Blue arrows depict the process of prototype calculation from feature embeddings and corresponding labels (masks), while red arrows indicate label prediction based on feature maps and prototypes. Background pixels are represented in black, foreground pixels in white, and low-confidence pixels in gray.
    }
\label{fig:method:overview}
\end{figure}

The term $\mathcal{P}^{q,c}$ denotes the query prototype for class $c$, derived using the pseudo-labeling method.
The final step involves performing the prediction based on the augmented support set, which now includes the enriched information from the consecutive query slices by adding query prototypes.
\begin{equation}
    \mathcal{P}^{aug} = \mathcal{P}^{s} \cup \mathcal{P}^{q}
\end{equation}
By leveraging the intrinsic information obtained from the query samples and expanding the support set through confidence-aware pseudo-labeling, our method aims to achieve superior segmentation performance in diverse settings and datasets.
Figure \ref{fig:method:overview} and Algorithm \ref{alg:inference} illustrate the overall procedure of the proposed method. As shown in Algorithm \ref{alg:inference}, the inference process can be divided into three stages. The first and third stages follow the traditional prototypical FSS approach, while the second stage represents the key contribution of our work, enhancing prediction accuracy in a plug-and-play manner.


\begin{algorithm}[!htb]
\caption{Few-shot Segmentation with Inference-Time Pseudo-Labeling}\label{alg:inference}
\begin{algorithmic}[1]
\State \textbf{Input:} Support set $\mathcal{S}$, Query set $\mathcal{Q}$, Model backbone $f_\theta$, Confidence threshold $\gamma$
\State \textbf{Output:} Final segmentation masks for query slices

\State \textbf{Stage 1: Initial Segmentation}
\For{each class $c$ in support set $\mathcal{S}$}
    \State Extract feature embeddings $f_\theta(x_k^c)$ for each support slice $x_k^c \in \mathcal{S}^c$
    \State Compute prototype $\mathcal{P}^{s,c}$ for class $c$ using masked average pooling
\EndFor
\For{each query slice $x^q_k \in \mathcal{Q}$}
    \State Compute feature embeddings $f_\theta(x^q_k)$
    \State Calculate softmax probabilities $p^{q,c}(h,w)$ for each pixel $(h, w)$
    using cosine similarity with prototypes $\mathcal{P}^{s,c}$
\EndFor

\State \textcolor{blue}{\textbf{Stage 2: Confidence-Aware Pseudo-Labeling}}
\textcolor{blue}{
\For{each query slice $x^q_k \in \mathcal{Q}$}
    \State Generate pseudo-labels $\hat{y}^q(h,w) = \operatorname*{argmax} p^{q,c}(h,w)$ for each pixel
    \For{each class $c$}
        \State Select confident regions where $p^{q,c}(h,w) \geq \gamma$
        \State Update query prototypes $\mathcal{P}^{q,c}$ for class $c$ from confident regions
    \EndFor
}
\State \textcolor{blue}{Augment the support prototypes with query prototypes: $\mathcal{P}^{aug} = \mathcal{P}^{s} \cup \mathcal{P}^{q}$}
\State \textbf{Stage 3: Final Segmentation}
\For{each query slice $x^q_k \in \mathcal{Q}$}
    \State Perform final segmentation using $\mathcal{P}^{aug}$ and query embeddings $f_\theta(x^q_k)$ 
\EndFor

\end{algorithmic}
\end{algorithm}
\vspace{3mm}
\section{Experiments}
\subsection{Experimental Setup}
In order to ensure consistency across experimental outcomes, we follow the evaluation guidelines in \cite{ouyang2020self}, including the use of consistent hyperparameters, data preprocessing methods, evaluation metrics, and comparison techniques. The network architecture, implementation, and training procedure closely follow the SSLALPNet \cite{ouyang2020self} approach. For inference, we employ a support volume containing three annotated slices (K = 3) to segment each query volume, consistent with the methodology explained in \cite{ouyang2020self}. Furthermore, to comprehensively assess our method's performance, we conduct tests under both settings introduced in \cite{ouyang2020self}. Setting 1 involves training the model in a self-supervised manner on all available slices across all scans, while Setting 2 involves partitioning the test label set into two groups—upper (spleen and liver) and lower abdomen (right and left kidney). In this setting, when testing on a specific group, all slices containing those organs will be excluded from the training set.

\textbf{Datasets} We conducted experiments on two widely-used medical datasets, specifically abdominal CT scans from the MICCAI 2015 MultiAtlas Abdomen Labeling Challenge \cite{landman2015miccai} and abdominal MRI scans from the ISBI 2019 Combined Healthy Abdominal Organ Segmentation Challenge \cite{kavur2021chaos}. Our experiments involved reporting average Dice scores based on 5-fold cross-validation. Following the previous studies \cite{ouyang2020self}, results of experiments are reported for four anatomical organs: the left kidney (LK), right kidney (RK), spleen, and liver.

\begin{table}[!htbp]
\caption{Dice score of different methods under setting 1; The training and testing datasets were derived from the same organ sets.} \vspace{5pt}
\label{table:setting1}
\centering
\resizebox{\textwidth}{!}{%
    \begin{tabular}{ccccccccccc}
        \hline
        \multirow{3}{*}{
            \diagbox[dir=NW,trim=rl]{Method~~~}{Dataset}
    } & \multicolumn{5}{c}{Abdominal-CT}                                                                        & \multicolumn{5}{c}{Abdominal-MRI}                                                         \\ \cline{2-6}  \cline{7-11} 
        & \multicolumn{2}{c}{Lower}       & \multicolumn{2}{c}{Upper}       & \multirow{2}{*}{Mean}               & \multicolumn{2}{c}{Lower}       & \multicolumn{2}{c}{Upper}       & \multirow{2}{*}{Mean} \\
        & LK             & RK             & Spleen         & Liver          &                                     & LK             & RK             & Spleen         & Liver          &                       \\ \hline
        SE-Net                   & 24.42          & 12.51          & 43.66          & 35.42          & \multicolumn{1}{c|}{29.00}          & 45.78          & 47.96          & 47.30          & 29.02          & 42.51                 \\
        Vanilla-PANet            & 20.67          & 21.19          & 36.04          & 49.55          & \multicolumn{1}{c|}{31.86}          & 30.99          & 32.19          & 40.58          & 50.40          & 38.53                 \\
        ALPNet                   & 29.12          & 31.32          & 41.00          & 65.07          & \multicolumn{1}{c|}{41.63}          & 44.73          & 48.42          & 49.61          & 62.35          & 51.28                 \\
        SSL-PANet                & 56.52          & 50.42          & 55.72          & 60.86          & \multicolumn{1}{c|}{57.88}          & 58.83          & 60.81          & 61.32          & 71.73          & 63.17                 \\
        SSL-ALPNet               & 72.36          & 71.81          & 70.96          & 78.29          & \multicolumn{1}{c|}{73.35}          & 81.92          & 85.18          & 72.18          & 76.10          & 78.84                 \\
        SSL-ALPNet + Ours        & \textbf{76.18} & \textbf{73.81} & \textbf{77.39} & \textbf{79.59} & \multicolumn{1}{c|}{\textbf{76.75}} & \textbf{82.11} & \textbf{87.15} & \textbf{74.16} & \textbf{78.43} & \textbf{80.46}        \\ \hline
    \end{tabular}%
    
}
\end{table}

\vspace{2mm}
\subsection{Results and Discussion}

\begin{figure}
  \centering
  \includegraphics[width=1\linewidth]{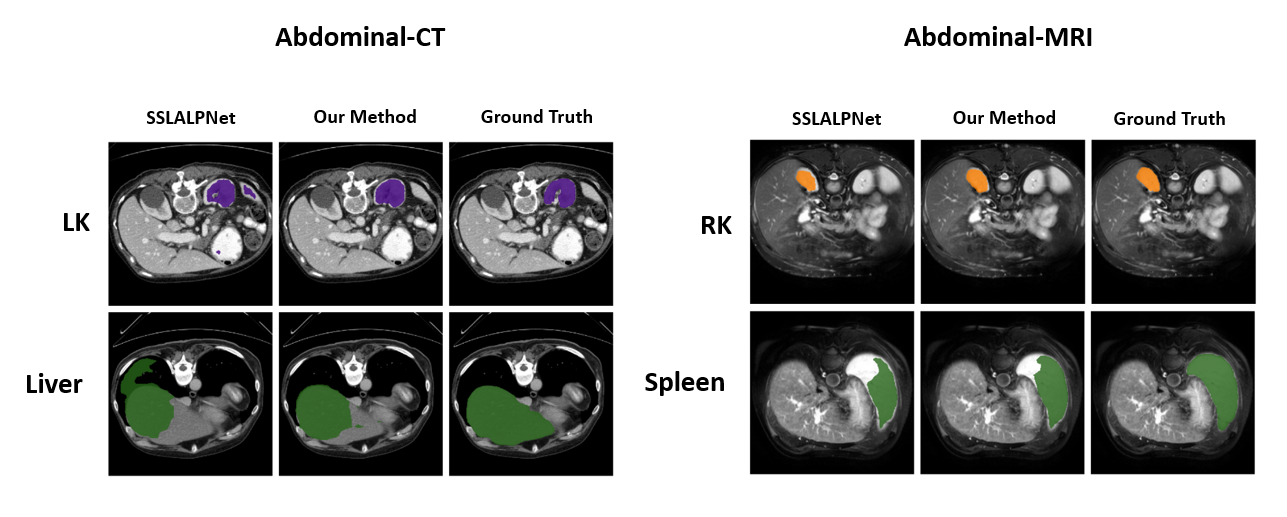}
   \caption{
   Qualitative comparison of our method with SSLALPNet \cite{ouyang2020self}}
\label{fig:qual}
\end{figure}

\textbf{Comparison with Other Methods} We conduct a comparative analysis involving PANet \cite{wang2019panet}, SSLALPNet \cite{ouyang2020self}, and ADNet \cite{hansen2022anomaly}. Our approach operates as a plug-in on top of SSLALPNet \cite{ouyang2020self}, serving as an inference strategy. Consequently, the training procedure and models remain identical to SSLALPNet. Nevertheless, through the utilization of our inference strategy leveraging query slices as unlabeled data, we have outperformed their results. Our method's effectiveness comes from the observation that slices within a scan tend to be more similar compared to the slices of different scans. This similarity enables the extracted prototypes from the query slices to be highly beneficial, as they closely resemble the query slice itself. Our approach has been evaluated under the conditions outlined in \cite{ouyang2020self}, with corresponding results provided in Tables \ref{table:setting1} and \ref{table:setting2}. Figure \ref{fig:qual} illustrates the effectiveness of our model qualitatively. Additionally, some ablation studies are conducted on fold 0 of the CT dataset under setting 1 which are detailed in the following sections. 

\begin{table}[!htbp]
    \caption{Dice score of different methods under setting 2; In this setting, when testing on a specific group, all slices containing those organs will be excluded from the training set.} \vspace{5pt}
    \label{table:setting2}
    \centering
    \resizebox{\textwidth}{!}{%
        \begin{tabular}{@{}ccccccccccc@{}}
            \toprule
            \multirow{3}{*}{
                \diagbox[dir=NW,trim=rl]{Method~~~~~~~~~}{Dataset}
        } & \multicolumn{5}{c}{Abdominal-CT}                                                                        & \multicolumn{5}{c}{Abdominal-MRI}                                                         \\ \cmidrule(l){2-6} \cmidrule(l){7-11}
            & \multicolumn{2}{c}{Lower}       & \multicolumn{2}{c}{Upper}       & \multirow{2}{*}{Mean}               & \multicolumn{2}{c}{Lower}       & \multicolumn{2}{c}{Upper}       & \multirow{2}{*}{Mean} \\
            & LK             & RK             & Spleen         & Liver          &                                     & LK             & RK             & Spleen         & Liver          &                       \\ \midrule
            SE-Net                  & 32.83          & 14.34          & 0.23           & 0.27           & \multicolumn{1}{c|}{11.91}          & 62.11          & 61.32          & 51.80          & 27.43          & 50.66                 \\
            Vanilla-PANet           & 32.34          & 17.37          & 29.59          & 38.42          & \multicolumn{1}{c|}{29.43}          & 53.45          & 38.64          & 50.90          & 42.26          & 46.33                 \\
            PANet                   & 37.58          & 34.69          & 43.73          & 61.71          & \multicolumn{1}{c|}{44.42}          & 47.71          & 47.95          & 58.73          & 64.99          & 54.85                 \\
            AD-Net                  & 63.84          & 56.98          & 61.84          & 73.95          & \multicolumn{1}{c|}{64.15}          & 71.89          & 76.02          & 65.84          & 76.03          & 72.20                 \\
            ALPNet-init             & 13.90          & 11.61          & 16.39          & 41.71          & \multicolumn{1}{c|}{20.90}          & 19.28          & 14.93          & 23.76          & 37.73          & 23.93                 \\
            ALPNet                  & 34.96          & 30.40          & 27.73          & 47.37          & \multicolumn{1}{c|}{35.11}          & 53.21          & 58.99          & 52.18          & 37.32          & 50.43                 \\
            SSL-PANet               & 37.58          & 34.69          & 43.73          & 61.71          & \multicolumn{1}{c|}{44.42}          & 47.71          & 47.95          & 58.73          & 64.99          & 54.85                 \\
            SSL-ALPNet              & 63.34          & 54.82          & 60.25          & 73.65          & \multicolumn{1}{c|}{63.02}          & 73.63          & 78.39          & 67.02          & 73.05          & 73.02                 \\
            Q-Net              & 63.26          & 58.37          & 63.36          & 74.36          & \multicolumn{1}{c|}{64.83}          & 74.05          & 77.52          & 67.43          & 78.71          & 74.43                 \\
            Cat-Net              & 63.36          & 60.05          & \textbf{67.65}          & \textbf{75.31}          & \multicolumn{1}{c|}{66.59}          & 74.01          & 78.90          & 68.83          & \textbf{78.98}          & 75.18                 \\
            SSL-ALPNet + Ours       & \textbf{69.02} & \textbf{62.03} & 63.95 & 75.07 & \multicolumn{1}{c|}{\textbf{67.52}} & \textbf{79.23} & \textbf{81.10} & \textbf{71.81} & 78.27 & \textbf{77.60}        \\ \bottomrule
        \end{tabular}%
    }
\end{table}

\vspace{2mm}
\textbf{Effect of window size} The segmentation process for a single query slice $i$ starts with the generation of pseudo-labels for all slices within the interval $[i-m, i+m]$, where $m$ represents a hyperparameter denoted as the window size. Subsequently, prototypes are extracted from these slices and employed for the segmentation of slice $i$. Empirical investigations are undertaken to find the optimal value for $m$ and the results are reported in Table \ref{table:ablationWindow}. When $m$ is set to zero, only the query prototypes of slice $i$ are utilized, which is deemed suboptimal due to potential inaccuracies in pseudo-labeling, resulting in prototypes of lower quality. By increasing the window size, the likelihood of encountering numerous inaccurate pseudo-labels decreases, consequently enhancing the quality of the prototype set and, by extension, improving segmentation outcomes. Conversely, an excessively large window size leads to an abundance of prototypes, thereby degrading the segmentation performance. Consequently, the optimal window size is neither small nor very large. Based on the findings, a window size of 7 emerges as optimal. 

\begin{table}[!htbp]
	\caption{Window sizes for employing prototypes extracted from queries, based on the target query's position within its respective volume.} \vspace{5pt}
	\label{table:ablationWindow}
	\centering
	\resizebox{0.77\textwidth}{!}{%
		\begin{tabular}{@{}c|ccccc@{}}
			\toprule
			Window Size & Left Kidney    & Right Kidney   & Spleen         & Liver          & Mean           \\ \midrule
			0           & 81.43          & 81.98          & 69.41          & 72.21          & 76.26          \\
			3           & 81.98          & 83.01          & 70.64          & \textbf{73.01} & 77.16          \\
			7           & 83.18          & 83.42          & \textbf{70.59} & 72.62          & \textbf{77.45} \\
			10          & \textbf{83.23} & \textbf{83.60} & 70.23          & 72.42          & 77.37          \\
			All Queries & 83.16          & 83.44          & 68.77          & 71.27          & 76.66          \\ \bottomrule
		\end{tabular}%
	}
\end{table}

\begin{table}[!htbp]
	\caption{Iterations to introduce confident pseudo-labels as additional prototypes.} \vspace{5pt}
	\label{table:ablationStages}
	\centering
	\resizebox{0.77\textwidth}{!}{%
		\begin{tabular}{@{}c|ccccc@{}}
			\toprule
			Num of Iterations           & Left Kidney    & Right Kidney   & Spleen         & Liver          & Mean           \\ \midrule
			\multicolumn{1}{c|}{2}  & \textbf{83.16} & \textbf{83.44} & \textbf{68.77} & \textbf{71.27} & \textbf{76.66} \\
			\multicolumn{1}{c|}{5}  & 82.17          & 82.05          & 61.98          & 70.23          & 74.11          \\
			\multicolumn{1}{c|}{8}  & 80.12          & 79.64          & 58.72          & 68.87          & 71.84          \\
			\multicolumn{1}{c|}{10} & 78.90          & 78.22          & 57.82          & 68.32          & 70.82          \\ \bottomrule
		\end{tabular}%
	}
\end{table}

\vspace{2mm}
\textbf{Number of stages} In our method we get the final segmentation mask in two iterations. Initially, the query volume undergoes segmentation, followed by the utilization of this segmentation as a pseudo-label in the next iteration. It is evident that this strategy can be iterated for multiple iterations. Nonetheless, according to 
Table \ref{table:ablationStages}, the optimal number of iterations is determined to be two, beyond which there is a noticeable decline in results. 

\begin{table}[!htbp]
	\caption{Strategies for constructing sets of prototypes.} \vspace{5pt}
	\label{table:ablationPrototypes}
	\centering
	\resizebox{0.85\textwidth}{!}{%
		\begin{tabular}{@{}c|ccccc@{}}
			\toprule
			Prototype Set & Left Kidney & Right Kidney & Spleen & Liver & Mean  \\ \midrule
			Only Support                                                                               & 82.20       & 76.19        & 68.62  & 70.11 & 74.03 \\
			\begin{tabular}[c]{@{}c@{}}Only Queries\end{tabular}           & 83.15       & 82.70        & 68.58  & 71.04 & 76.37 \\
			\begin{tabular}[c]{@{}c@{}}Support and Queries\end{tabular} & \textbf{83.16}       & \textbf{83.44}        & \textbf{68.77}  & \textbf{71.27} & \textbf{76.66} \\ \bottomrule
		\end{tabular}%
	}
\end{table}

\vspace{2mm}
\textbf{Prototype set construction} Upon generating pseudo-labels for queries and extracting prototypes from these queries, we have several approaches to segmenting the query slices using the prototypes from both the queries and the support. Initially, we may opt for utilizing solely the support prototypes (extracted in Stage 1), alternatively, rely exclusively on the query prototypes, or, in another strategy, employ a mix of both. Through experimental analysis, we determine the effectiveness of each method, with the findings detailed in Table \ref{table:ablationPrototypes}. The outcomes indicate that forming an augmented set of prototypes, incorporating both support and query prototypes proves to be the most effective tactic.


\section{Conclusion}
This study presents a novel three-stage inference-time method for few-shot segmentation (FSS), introducing a confidence-aware pseudo-labeling process to refine predictions. The proposed approach modifies only the inference phase, keeping the training procedure intact, which makes it a plug-and-play enhancement to existing FSS frameworks. In the first stage, initial segmentation is performed using prototypes derived from the support set. In the second stage, the confidence-aware pseudo-labeling mechanism identifies reliable regions in the query slices, which are then used to update the query prototypes. Finally, the augmented prototype set, combining support and query information, is employed for final segmentation.

Our approach effectively leverages the information from unlabeled query samples to improve segmentation performance without additional supervision. By focusing on inference-time optimization, this method avoids the complexities of retraining and shows potential for practical application across a variety of FSS tasks. Empirical results demonstrate that this technique yields enhanced segmentation accuracy, particularly in scenarios where labeled support data is scarce, underlining the significance of incorporating query data into the prototype refinement process.

\newpage

\bibliography{references}

\end{document}